# Prediction by Machine Learning Analysis of Genomic Data Phenotypic Frost Tolerance in Perccottus glenii

Lilin Fan[1,2,3,*], Xuqing Chai[1,2,3,*], Zhixiong Tian[1(✉)], Yihang Qiao[1], Zhen Wang[1], and Yifan Zhang[1]

[1] School of Computer Information and Engineering, Henan Normal University, Xinxiang 453000, China
[2] Henan Engineering Laboratory of Smart Business and Internet of Things Technology
[3] Henan Normal University High Performance Computing Center
2208283040@stu.htu.edu.cn

**Abstract.** Analyzing the genome sequence of the only fish species known to possess freezing tolerance, Perccottus glenii, is crucial for understanding how organisms adapt to extreme environments. Traditional biological analysis methods are time-consuming and have limited accuracy. Therefore, we will employ machine learning techniques to analyze the genomic sequences of Perccottus glenii, using Neodontobutis hainanens as a comparison group to identify differential genes. Firstly, we propose five gene sequence vectorization methods and a method for handling super-long gene sequences, and we compare three vectorization methods: sequential encoding, One-Hot encoding, and K-mer encoding, to identify the optimal encoding method. Secondly, we construct four classification models: random forest, LightBGM, XGBoost, and decision tree. The dataset used for these models is derived from the NCBI database. We determine the optimal K-value for the K-mer encoding method, and extract and vectorize the gene sequence matrix, with the best-performing model, random forest, achieving a classification accuracy of up to 99.98%. Finally, we use SHAP values to perform interpretability analysis on the classification models, and through tenfold cross-validation and AUC metrics, we identify the top 10 features that contribute the most to the model's classification accuracy, thereby recognizing the differential genes for freezing tolerance phenotype in Perccottus glenii. We validate the identified differential genes using the biological software BLAST. Conclusion: Our study demonstrates that machine learning methods can replace traditional manual methods for identifying differential genes associated with freezing tolerance phenotype in Perccottus glenii.

* These authors contributed equally to this work and should be considered co-first authors.

## 1      Introduction

Currently, Perccottus glenii is the only fish known to possess freezing tolerance, while its closely related species Neodontobutis hainanens does not have the ability to withstand cold temperatures. Analyzing the genomic sequences of these two species to identify the phenotypic genes that contribute to the cold tolerance of Perccottus glenii is of great significance for understanding how organisms adapt to extreme environments [1].

Through bioinformatics analysis of genomic differences between the two species, scientists can gain insights into the mechanisms underlying the cold tolerance of Perccottus glenii, providing valuable genetic resources for future research. However, bioinformatics analysis has its limitations. For instance, with the advancement of sequencing technologies [2,3], the volume of gene sequence data is continually increasing, and the analysis of these large datasets requires efficient computational power and storage capacity. Additionally, bioinformatics analysis has limited accuracy when it comes to the alignment of large or complex genomes. Overcoming these limitations presents a new challenge. Traditional experimental methods for bioinformatics analysis of the genomes of these two fish species are time-consuming and labor-intensive, and they are limited by the experience and skills of the experimenters. The manual operation involved in the analysis process can lead to errors and subjective judgments, resulting in limited accuracy for traditional bioinformatics methods in genome alignment. To effectively address these issues, efforts are being made to apply machine learning methods [4] to the analysis of genomic sequences of organisms, with interdisciplinary collaboration aimed at solving the current problems faced [5].

This article fully leverages the strengths of bioinformatics and computer science, combining traditional biological gene sequence analysis with machine learning methods to collaboratively process large-scale biological gene sequence data [6], automating the gene sequence analysis process. This transformation makes the originally cumbersome and time-consuming analysis simple and efficient, reducing manual intervention while improving analytical efficiency. Through machine learning methods, patterns and features within gene sequences can be accurately identified [7], achieving higher prediction accuracy and further uncovering key information within gene sequences compared to traditional methods.

The main contributions of this paper include the use of machine learning methods to replace traditional manual methods for the analysis of genomes from different species, reducing workload and human error, and enhancing the efficiency and accuracy of genomic analysis. Our specific contributions include:

1. Data Analysis: We propose five methods for gene sequence vectorization and conduct a comparative analysis of three of these methods to assess their feasibility.
2. Modeling: We employ four different machine learning models to learn features from gene sequences and compare their classification effects to evaluate the feasibility of using machine learning for genome comparison.
3. Biological Interpretability: We perform interpretability analyses on the machine learning models, correlating the learned features with their biological significance to explore the potential for machine learning features to enhance biological interpretability.

All the code used in this paper is publicly available at https://github.com/TZX888/gene-sequence.git for future reference and replication.

## 2 Materials and Methods

### 2.1 Gene Sequence Encoding Methods

In order to use machine learning methods on gene sequence datasets, the following five gene sequence vectorization methods are proposed [8].

**Sequential encoding**

In machine learning, Sequential Encoding [9] is a method that maps the individual categories of a discrete feature to integer ordinal numbers. This approach is suitable for ordered features where there is a sequential relationship between categories but no explicit quantitative meaning. As illustrated in Fig. 1, ordinal encoding assigns integers to categories according to their order.

Sequential DNA sequence coding is a method proposed by Allen Chieng, Hoon Choong, and Nung Kion Lee in their paper [10]. This method encodes nucleotide sequences into numerical values, with A represented by 0.25, C by 0.5, G by 0.75, and T by 1.00. Any other base is represented by 0.0. While this coding scheme was experimentally found to be effective, the significance of the specific numerical choices for this encoding is not readily apparent.

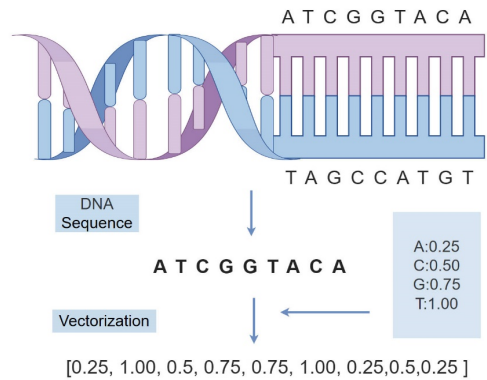

**Fig. 1.** Sequential Coding Schematic

**One-Hot Encoding**

One-Hot Encoding [11] is one of the most commonly used methods for encoding categorical features in machine learning. It works by creating a new binary feature for each category of a discrete attribute. For each sample, only one of these binary features is set to 1, indicating the presence of the corresponding category, while all other features are set to 0. This method is particularly suitable for attributes with a finite number of categories. As shown in Fig. 2, an example of One-Hot Encoding encodes the nucleotides ATCG as [0,0,0,1], [0,0,1,0], [0,1,0,0], and [1,0,0,0], respectively. A sequence of 1000bp would thus be transformed into a matrix of size 1000*4. Since One-Hot Encoding converts categories into discrete binary vectors, it does not preserve the ordinal information between categories. This means that when decoding, it is necessary to know the specific meaning of each binary bit, which can be less intuitive in the context of DNA sequence encoding.

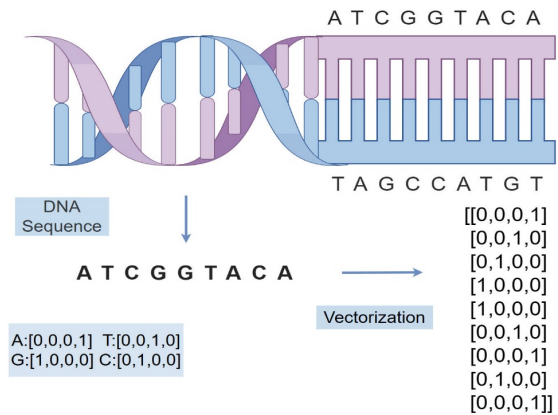

**Fig. 2.** One-Hot Encoding Schematic

## K-mer Encoding

The methods of One-hot encoding and sequential encoding cannot produce vectors of consistent length, as they achieve vector uniformity by truncating sequences or padding with zeros. K-mer encoding [12] is a method that transforms genomic sequences into vectors by breaking the sequence into overlapping k-mer fragments and calculating the frequency of each k-mer, treating the genomic sequence as natural language processing. As shown in Fig. 3, the sequence S1 is split by sliding with K=3. The sequence ATCGCA can be divided into ATC, TCG, CGC, GCA. After performing the same operation on the remaining n-1 sequences, the n sequences are segmented to form a set of sequences with a length of 3, denoted as N. Taking the union of N forms a gene sequence dictionary with unique IDs. In Fig. 3, the gene dictionary is used as the columns of a matrix, with the n sequences as the rows, to construct a frequency matrix.[13] The matrix calculates the number of occurrences of each sequence's gene fragment in the gene dictionary. K-mer encoding generates a machine learning feature matrix through the above method.

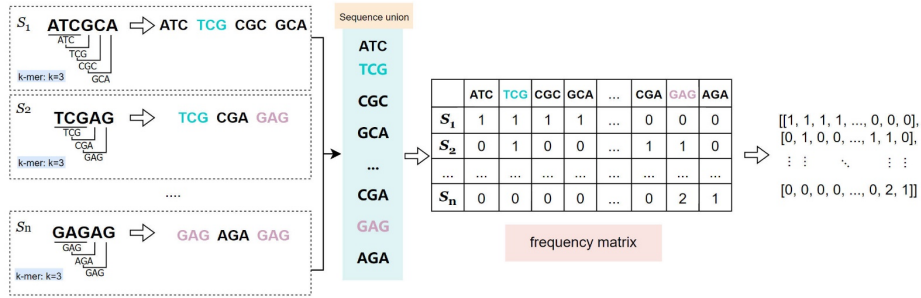

**Fig. 3.** k-mer coding schematic

## Figure Embedding

The vectorized representation of gene sequences utilizes node-to-node co-occurrence relationships in a graph, preserving the sequence information within the graph's data structure through the k-mer method's continuity. Fig. 4 illustrates the entire process of k-mer coding with K=3 as an example, demonstrating how gene sequences are represented in a graphical format. In the graph, the DNA sequence GTCGACGAC is sequentially segmented into neighboring 3-mers, and these neighboring 3-mers are connected to each other. The nodes in the feature extraction graph represent the segments after 3-mer segmentation, the edges represent the continuity between the nodes, and the weights of the edges between the nodes represent the frequency with which neighboring nodes appear consecutively. Once the DNA sequence is represented as a graph using the aforementioned method, the feature graph can be transformed into a vector based on graph methods and subsequently learned using graph neural network techniques.

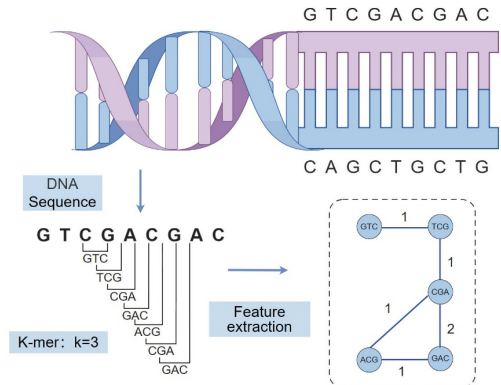

**Fig. 4.** Coding Schematic

## Picture channel coding DNA sequences

The image channel, in RGB color mode, refers to the separate red, green, and blue components. A complete image is composed of three channels: red, green, and blue, which work together to create the full image. The genome sequence can be represented as an image using the image channel approach, where the sequence is treated as an image consisting of four channels (A, T, C, and G). As shown in Fig. 5, the DNA sequences S1, S2, S3 are mapped sequentially into the four channels (A, T, C, G) in pixel order. Each DNA sequence can thus be transformed into a four-channel image, allowing the prediction problem of DNA sequences to be converted into a classification problem for binary images.

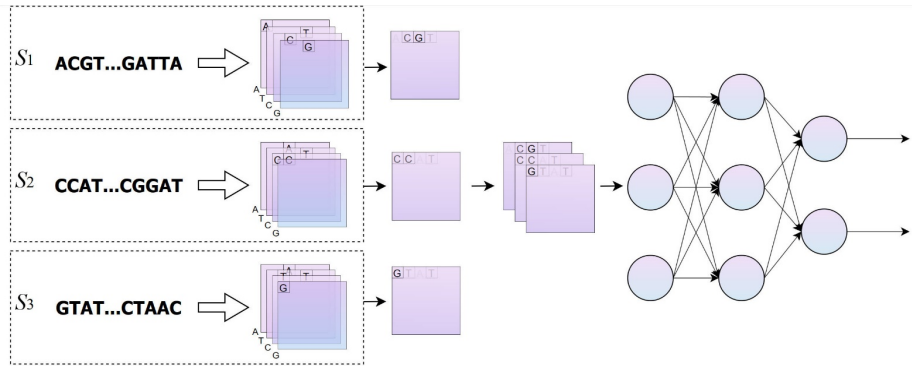

**Fig. 5.** Picture Channel Encoding DNA Sequence Schematic

The above article proposes five different methods for DNA vectorization, and the classification efficacy of these methods will be validated in subsequent studies. When dealing with ultra-long gene sequences (spanning billions of base pairs), a key challenge in DNA sequence vectorization methods is how to effectively process them. A common strategy in machine learning for handling such sequences is truncation. While truncation offers the advantage of reducing the sequence to a more manageable length, a significant disadvantage is the potential loss of valuable data that could be crucial for accurate predictions. Another approach is to represent ultra-long sequences hierarchically, which aids the model in handling extended sequence data. The core principle of this strategy involves breaking down the ultra-long sequences into several levels, with each level processing a segment of the sequence. This reduces the sequence length that the model needs to handle while preserving the sequence's hierarchical information. Additionally, processing ultra-long gene sequences with a Convolutional Neural Network (CNN) can capture local features through sliding convolutional kernels and reduce sequence length using pooling layers during data processing. By stacking multiple convolutional layers, a CNN can manage ultra-long sequences. However, these methods may suffer from issues such as information loss and a lack of biological interpretability. To address these concerns, this paper combines the K-mer coding method to divide the ultra-long sequences into multiple fixed-length sub-sequences and constructs a gene vector dictionary to train each sub-sequence as an independent feature. This approach can prevent the loss of gene information, and the experimental results are biologically interpretable.

## 2.2    Hardware Configuration

The entire experimental hardware configuration is as follows: We used two servers with the same specifications, each equipped with two AMD EPYC 9654 96-Core Processors, totaling 192 cores per server. Each server has a memory capacity of 2TB.

## 2.3    Dataset

For the gene sequence data of Perccottus glenii and Neodontobutis hainanens, this paper downloaded the sequence data of the two fish from the NCBI BioProject database under the accession numbers PRJNA818152 (P.glenii) and PRJNA818180 (N.hainanensis), and Perccottus glenii was used as the study subject and Neodontobutis hainanens was used as the control.

The format of the downloaded dataset is the FASTA format (see Fig. 6), which is an ASCII text-based format that can store one or more nucleotide sequences or peptide sequence data. In the FASTA format, each sequence data begins with a single line of description, followed immediately by one or more lines of sequence data. The next sequence data does the same, and the cycle continues.

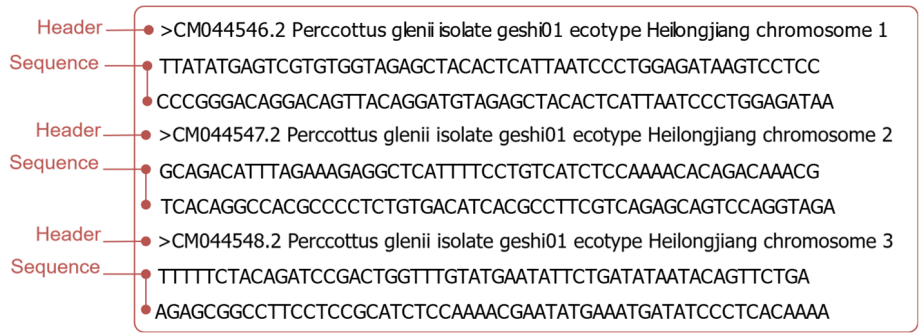

**Fig. 6.** FASTA format

The gene sequence files of Perodontobottus glenii (Gfish) and gene sequence data of Neodontobottus hainanens (Hfish) were analyzed, and in Fig. 7, the variable X represents the number of gene sequences, and the variable y denotes the length of gene sequences, and it can be seen that, most of the gene lengths of each genome fluctuate within a certain interval, and a small portion of the lengths of the gene sequences belongs to the outlier points.

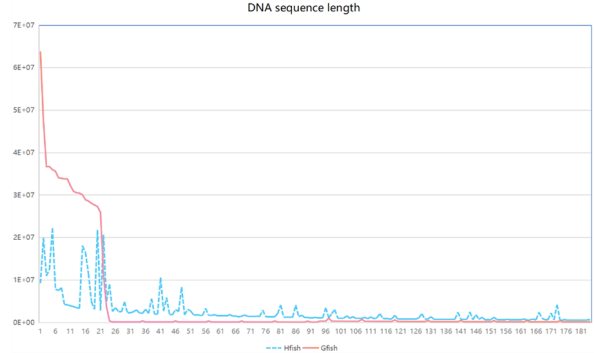

**Fig. 7.** Hfish and Gfish DNA sequence length

Since the Perccottus glenii genes for frost resistance accounted for only a small portion of the dataset, the sample size of the two categories was unbalanced. The commonly used methods to deal with the unbalanced dataset include under-sampling, over-sampling and other algorithms, in this paper, we adopt Synthetic minorty over-sampling technique (SMOTE algorithm[16]) which has good performance and is widely used.The basic idea of SMOTE algorithm is to analyze the minorty samples and add new samples to the dataset according to the minorty samples artificially, so as to increase the number of samples, and then achieve the overall sample size, thus to achieve the overall dataset. The basic idea of the SMOTE algorithm is to analyze the minority samples and artificially synthesize new samples based on the minority samples to be added to the dataset, so as to increase the number of samples, and then achieve the sample balance of the overall dataset [18].

### 2.4    K-mer Encoding K-value Selection

When performing K-mer encoding on genomic sequence data, the choice of the K-value has a significant impact on accuracy. Ideally, K should be large enough for k-mers to map to unique positions in the genome. However, an excessively large K can reduce the probability of removing erroneous bases represented by low-frequency k-mers (increasing the error rate), decrease k-mer depth (resulting in a less pronounced peak in the k-mer frequency distribution), and increase the usage of computational resources. Therefore, it is necessary to select an appropriate K-value to balance computational complexity and the integrity of sequence information. We used KmerGenie to analyze the Gfish and Hfish genomes. Fig. 8 shows the frequency distribution of each k-mer. KmerGenie assesses the highest k-value for the total size of the genome as the "optimal k-mer", providing a reference for the subsequent k-value to be used for genomic sequence encoding with this sequencing data. From Fig. 8(a), it can be observed that the optimal K for the Gfish genome is 19. From Fig. 8(b), it can be seen that the optimal K for the Hfish genome is 25, with 19 being the next best choice. Considering computational resource constraints, we selected K=19 for K-mer encoding.

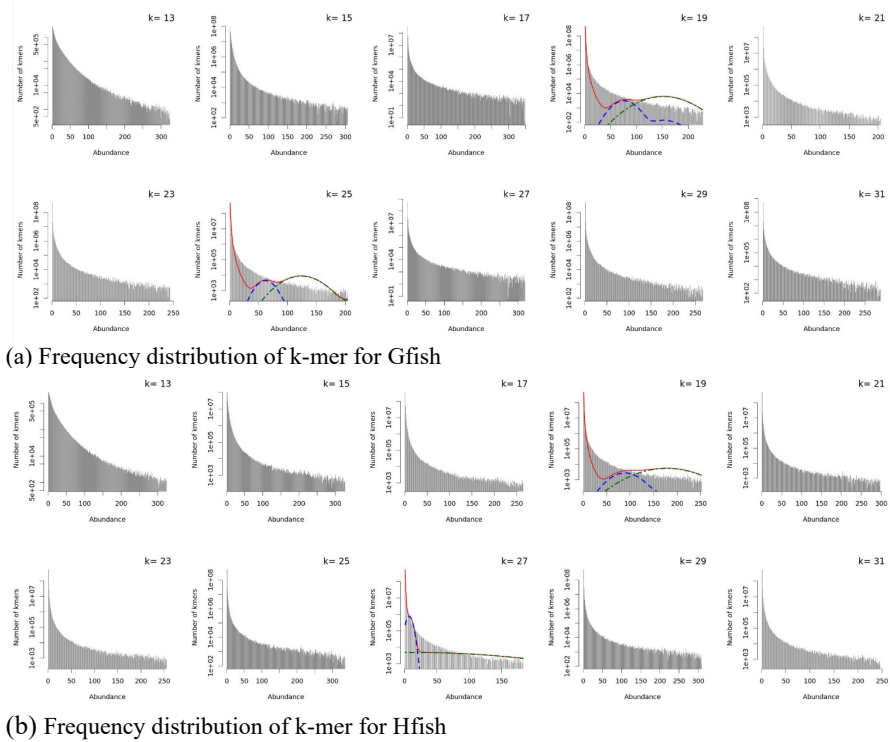

(a) Frequency distribution of k-mer for Gfish

(b) Frequency distribution of k-mer for Hfish

**Fig. 8.** From top to bottom, (a) shows the k-mer frequency distribution of the Perodonto-bottus glenii, with the optimal K=19; (b) shows the k-mer frequency distribution of the Neodonto-bottus hainanens, with the optimal K=27. The red curve represents the observed k-mer curve; the blue curve represents the heterozygous k-mer curve; and the green curve represents the homozygous k-mer curve.

## 2.5   Comparative Experiment

The genome sequence $s = N_1 N_2 \cdots N_L$, $N_i \in \{A, T, C, G\}$, $i = 1, 2, \cdots, L$, and $L$ is the length of the sequence. Construct the dictionary $E = \{N_1 N_2 \cdots N_k\}$, $N_i \in \{A, T, C, G\}$, Dictionary $E$ contains $k$ sub-elements of 4. If every two neighboring sites of the sequence are scanned consecutively from the beginning with k=19, e.g., site pairs $(1,19), (2,20), \cdots, (L-19, L)$, from which the $L - 19$ feature sequences converted from the original sequences can be obtained [14]. The feature sequence segmentation is performed on all gene sequences, and the dictionary $E$ is the set of full alignments of gene sequences of A,T,C,G [15].

The 7970 Perccottus glenii gene sequences and 184 Neodontobutis hainanens gene sequences after segmentation were processed and randomly divided into a training set (5708 sequences) and a test set (2446 sequences) in a 7:3 ratio.Different population category genes were tagged manually for supervised learning, and the gene sequences were vectorized using sequential coding, Onehot coding, and K-mer coding, respectively, to make a side-by-side comparison of the three gene coding methods.

To investigate whether machine learning methods are suitable for stratifying the freezing tolerance phenotype of Gobioides broussonnetii and its homologous species Perccottus glenii, we selected the optimal encoding method from the three aforementioned encoding methods. We constructed four different models—Random Forest[19], LightGBM[20], XGBoost[21], and Decision Trees[22]—for comparative classification testing. Additionally, we conducted parameter optimization to select the model with the best predictive performance for subsequent research.

## 2.6    Evaluation and Interpretation of Models

In order to validate the model, this study uses 10-fold cross-validation[23] to judge the model. In 10-fold cross-validation, the dataset is split into 10 different subsets, and 9 are used as the training set each time, and the remaining 1 is used as the test set to validate the model performance, and finally, the average performance of the 10 model performances is obtained as the result of the 10-fold cross-validation. In addition, we also used the learning curve during[24] training to evaluate the model. Finally, the model with the best classification result was interpreted by SHAP value[25] and the importance ranking of each gene feature was obtained, and then the gene sequences were sequentially included in the model and their area under the curve (AUC)[26] was calculated according to the importance ranking of the features, and a suitable number of gene segments were selected for subsequent studies considering their stability.

To determine which features are important in stratifying the frost tolerance phenotype of Perccottus glenii, this paper uses SHAP to interpret the trained model described above.SHAP (SHapley Additive exPlanations[27]) is a method for elucidating the underlying mechanisms of machine learning model predictions, which is based on the Shapley value theory from game theory. It quantifies the individual impact of each feature on the model's predictive outcomes and provides an analytical framework that can explain both the overall predictive behavior of the model and the influence of features on individual prediction instances simultaneously. This technique was first proposed by Lundberg and Lee in 2017. SHAP is utilized to enhance the interpretability of machine learning models by treating each feature as a contributor to the model's predictions. It calculates the SHAP values for each feature through perturbation, determining the degree of contribution and thereby interpreting the model. For a single prediction instance, the calculation of the SHAP value can be approximated by Equation (1):

$$\emptyset_i = \sum_{S \subseteq F \setminus \{i\}} \frac{|S|!(|F|-|S|-1)!}{|F|!} [f_s(x_s) - f_s(x_s \cup \{x_i\})] \tag{1}$$

Where $\emptyset_i$ is the SHAP value for feature $i$, $F$ is the set of all features, $S$ is a subset of $F$ that does not include feature $i$, $x_s$, represents the feature values in set $S$, $f_s$ is the model's prediction on feature set $S$, $|S|$ is the number of features in set $S$, ! denotes factorial.

And the entire framework of the SHAP method is depicted in Fig. 9.Using SHAP values to interpret the predictions of the test set, features with high SHAP values make significant contributions to the classification predictions. Such features are primarily associated with differential genes between the two fish species. The research objective is transformed into identifying meaningful biological features through those with high SHAP values. In this paper, SHAP is used to interpret the model with the best classification performance. The features are ranked from high to low based on their SHAP values, and the corresponding gene sequences are identified for verification.

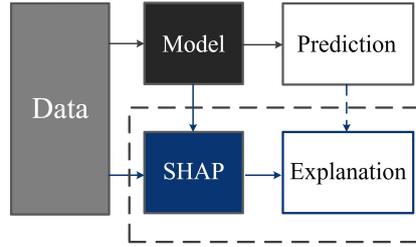

**Fig. 9.** SHAP framework diagram

### 2.7 Evaluation Indicators

To evaluate the classification performance of the model, metrics such as accuracy, precision, recall, and F1-score, which are commonly used in classification models, are employed. These metrics are calculated using TP (True Positive), FP (False Positive), FN (False Negative), and TN (True Negative). Specifically, TP represents the number of instances that are correctly classified as positive in both the true and predicted categories, FP represents the number of instances that are incorrectly classified as positive in the predicted category when they are actually negative in the true category, FN represents the number of instances that are incorrectly classified as negative in the predicted category when they are actually positive in the true category, and TN represents the number of instances that are correctly classified as negative in both the true and predicted categories.

Accuracy measures the proportion of correct classifications, as shown in equation (2):

$$Accuracy = \frac{TP+TN}{TP+FP+FN+TN} \qquad (2)$$

The precision refers to how many of all samples judged positive by the model are truly positive, as shown in equation (3):

$$Precision = \frac{TP}{TP+FP} \qquad (3)$$

Recall refers to how many of all positive samples are judged positive by the model, as shown in equation (4):

$$Recall = \frac{TP}{TP+FN} \qquad (4)$$

The F1-score [28] is the reconciled average of precision and recall as shown in equation (5):

$$F1 - score = 2 \times \frac{Precision \times Recall}{Precision + Recall} \tag{5}$$

## 3    Results and Analysis

### 3.1    K-mer Encoding Leads to Better Performance

In the comparative experiment of the three vectorization methods, we utilized gene sequences of the species Perccottus glenii and Neodontobutis hainanens as the dataset. The gene sequences were vectorized using three methods: ordinal encoding, One-Hot encoding, and K-mer encoding, which were transformed into input matrices for machine learning. Random Forest was selected as the learner to compare the effects of the three encoding methods. The confusion matrices for the three encoding methods are shown in Fig. 10. Fig. 10(a) presents the confusion matrix results for ordinal encoding input into Random Forest, Fig. 10(b) for One-Hot encoding, and Fig. 10(c) for K-mer encoding. From Fig. 10, it can be observed that among the three encoding methods, K-mer encoding yields the best classification performance. As indicated in Table 1, when using K-mer encoding as input in the Random Forest model, the accuracy, precision, recall, and F1-score are 0.9998, 0.9998, 0.9998, and 0.9998, respectively, demonstrating the superiority of the K-mer encoding method over ordinal encoding and One-Hot encoding. Subsequent experiments will therefore employ K-mer encoding.

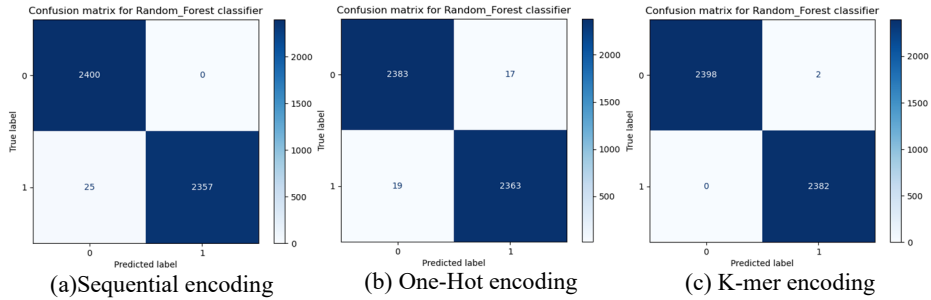

(a)Sequential encoding    (b) One-Hot encoding    (c) K-mer encoding

**Fig. 10.** From left to right, (a) depicts the confusion matrix for the results of the random forest model with sequential encoding input, (b) shows the confusion matrix for the results of the random forest model with One-Hot encoding input, (c) presents the confusion matrix for the results of the random forest model with K-mer encoding input.

**Table 1.** Comparison chart of the classification results of the three coding methods

| Methods | tarfets | model | Accuracy | Precision | Recall | F1-score |
|---------|---------|-------|----------|-----------|--------|----------|
| Sequential encoding | | | 0.9948 | 0.9948 | 0.9948 | 0.9948 |
| One-Hot encoding | Testset | Random Forest | 0.9925 | 0.9925 | 0.9925 | 0.9925 |
| **K-mer encoding** | | | **0.9998** | **0.9998** | **0.9998** | **0.9998** |

## 3.2    Random Forest Models Lead to Better Classification Performance

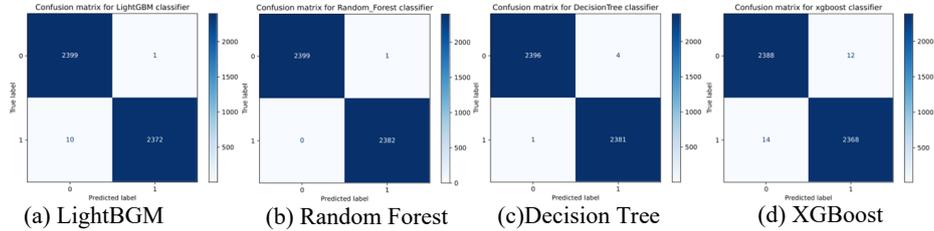

(a) LightBGM    (b) Random Forest    (c)Decision Tree    (d) XGBoost

**Fig. 11.** From left to right, (a) is the confusion matrix for the classification results of the LightBGM model, (b) is the confusion matrix for the Random Forest model, (c) is the confusion matrix for the Decision Tree model, and (d) is the confusion matrix for the XGBoost model.

**Table 2.** Comparison of classification results of four different models

| Model | Example | Accuracy | Precision | Recall | F1-scorce |
|---|---|---|---|---|---|
| XGBoost | Testset | 0.9946 | 0.9946 | 0.9946 | 0.9946 |
| | ten-fold cross validation | 0.9490 | 0.9510 | 0.9490 | 0.9490 |
| Decision Tree | Testset | 0.9990 | 0.9990 | 0.9990 | 0.9990 |
| | ten-fold cross validation | 0.9389 | 0.9388 | 0.9389 | 0.9389 |
| LightGBM | Testset | 0.9977 | 0.9977 | 0.9977 | 0.9977 |
| | ten-fold cross validation | 0.9579 | 0.9580 | 0.9579 | 0.9579 |
| Random Forest | Testset | **0.9998** | **0.9998** | **0.9998** | **0.9998** |
| | ten-fold cross validation | 0.9489 | 0.9515 | 0.9492 | 0.9488 |

In the model comparison experiment, based on the K-mer vectorization method, we constructed four different models on the training set: Random Forest, LightGBM, XGBoost, and Decision Tree. The confusion matrices for the classification results of these models are shown in Fig. 11.Fig. 11(a) depicts the confusion matrix for the classification results of the LightGBM model, Fig. 11(b) shows the confusion matrix for the Random Forest model, Fig. 11(c) presents the confusion matrix for the Decision Tree model, and Fig. 11(d) displays the confusion matrix for the XGBoost model. To evaluate the performance of these four models, we employed various evaluation metrics. The areas under the ROC curves for the Random Forest, LightGBM, XGBoost, and Decision Tree models constructed on the test set were 0.9996, 0.9994, 0.9991, and 0.9879, respectively. Additionally, as shown in Table 2, the table displays the accuracy, precision, recall, and F1-scores for the four models, along with the results of 10-fold cross-validation. The results indicate that the Random Forest model achieved accuracy, precision, recall, and F1-scores of 0.9998, 0.9998, 0.9998, and 0.9998, respectively, on the dataset. The results of the 10-fold cross-validation were 0.9489, 0.9515, 0.9492, and 0.9488, respectively.Subsequently, we conducted a learning curve test on the Random Forest model to verify whether the model was overfitting[29]. As shown in Fig. 12, the training score remained at a relatively high level, while the validation score gradually increased. Although the training set score was close to 1, the cross-validation score gradually increased, with the last two points almost coinciding, indicating that there was no overfitting at this point. We then further optimized the Random Forest model and continued with subsequent analyses.

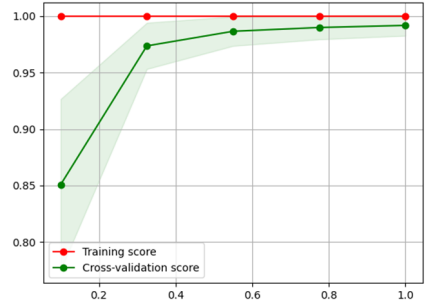

**Fig. 12.** Learning Curve for Random Forest Models

### 3.3    Interpretable Analysis

In order to identify the features that make significant contributions to the model's classification results, we employed SHAP values to interpret the model. Fig. 14 illustrates the ranking of the top 9 features in terms of their importance to the model, with the features' importance to the model's classification outcomes arranged from highest to lowest. The ranking order is determined by the average absolute SHAP value of a feature within the model. Subsequently, we sequentially added features to the model according to this ranking and retrained it, evaluating the model each time with AUC. The results revealed that regardless of whether it was in the validation set or cross-validation, the AUC reached a relatively stable state when the 9th gene segment was included as a feature (see Fig. 13). In line with Occam's Razor principle[30], we ultimately selected the gene segments corresponding to the top 9 ranked features for further research.

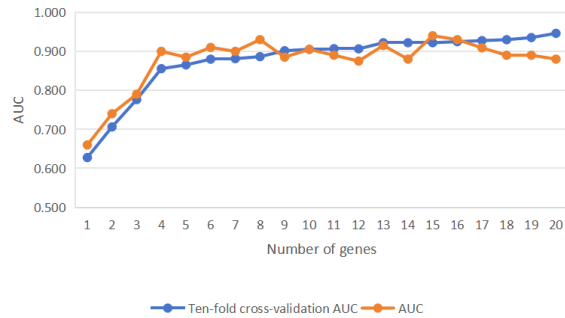

**Fig. 13.** Feature Selection Diagram

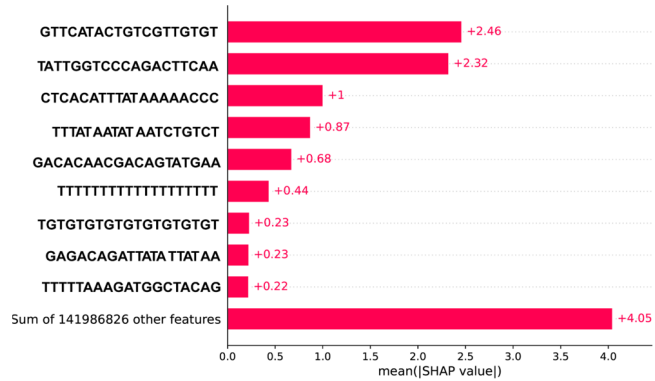

**Fig. 14.** SHAP Values in Random Forest Models

### 3.4 Discussion of Results

In this study on the freezing tolerance phenotype of Perccottus glenii based on machine learning methods, we first proposed five gene sequence vectorization methods and a method for handling super-long gene sequences. We then conducted experiments to compare the classification effects of ordinal encoding, One-Hot encoding, and K-mer encoding, demonstrating that the K-mer method outperforms the other two encoding methods. Subsequently, using the K-mer encoding method, we applied four different machine learning models—random forest, LightGBM, XGBoost, and decision tree—to classify the DNA sequences of the two species. We compared the performance of these four models in terms of Accuracy, Precision, Recall, and F1-score on both the test set and through 10-fold cross-validation, showing that the random forest model is superior to the other three models. Finally, we used SHAP values to interpret the random forest model and successfully identified 9 differential genes significantly associated with the freezing tolerance phenotype for further research. To verify the correctness of the results, we used the biological software BLAST (Basic Local Alignment Search Tool) to align the 9 differential genes to their corresponding positions in the genome, confirming the feasibility of identifying differential genes based on machine learning. The gene vectorization methods and the handling method for super-long gene sequences proposed in this paper offer new insights for gene analysis using machine learning.

## 4 Conclusion

In this work, we explored and compared different gene encoding methods that could affect the performance of classification models. The results indicated that the K-mer encoding method outperforms the other encoding methods tested. We further compared the classification performance of different models under K-mer encoding and found that the random forest classification model had the best performance. Using SHAP val-ues, we effectively interpreted the model features and identified the Top 9 features corresponding to differential genes. Finally, we aligned the identified differential genes and found their corresponding gene loci in the genome, demonstrating the effectiveness of machine learning methods for the identification of differential genes.

Currently, traditional manual methods for identifying differential genes have several limitations, including being time-consuming and labor-intensive, subjective, difficult to handle large-scale data, lacking automation, and being limited by prior knowledge. In response to these issues, this paper proposes a new method for identifying differential genes based on machine learning. By extracting gene sequence features from different species using various methods, establishing gene classification models, and conducting interpretable analyses, we were able to identify differential genes between species. We have proven the effectiveness and practicality of the proposed method in identifying differential genes. In the future, we expect this method to be widely applied in genomics research and to provide new insights into biodiversity, evolutionary relationships, and disease mechanisms.

**Acknowledgments.** This work was supported by the National Natural Science Foundation of China (Grant No. 12274117), Project supported by the China University Industry University Research Innovation Fund - New Generation Information Technology Innovation Project( 2020ITA07040), the Program for Innovative Research Team (in Science and Technology) in University of Henan Province (Grant No. 24IRTSTHN025), the Young Top-notch Talents Project of Henan Province (2021 year), the High Performance Computing Centre of Henan Normal University, and  the Supercomputing Center of University of Science and Technology of China.

**Disclosure of Interests.** The authors have no competing interests to declare that are relevant to the content of this article.